\documentclass{article}

\PassOptionsToPackage{numbers, compress}{natbib}

\usepackage[preprint]{neurips_2020}

\usepackage[utf8]{inputenc} 
\usepackage[T1]{fontenc}    
\usepackage{hyperref}       
\usepackage{url}            
\usepackage{booktabs}       
\usepackage{amsfonts}       
\usepackage{nicefrac}       
\usepackage{microtype}      
\usepackage{graphicx}
\usepackage{wrapfig}
\usepackage{multirow}
\usepackage[linesnumbered,ruled,vlined,linesnumbered]{algorithm2e}
\usepackage{makecell}

\newcommand{\etal}{\textit{et al}. }
\newcommand{\ie}{\textit{i}.\textit{e}. }
\newcommand{\eg}{\textit{e}.\textit{g}. }

\SetKwInput{KwInput}{Input}                
\SetKwInput{KwOutput}{Output}              

\title{Learning High-Resolution Domain-Specific Representations with a GAN Generator}

\author{%
  Danil Galeev \\
  Samsung AI \\
  \texttt{d.galeev@samsung.com} \\
  \And
  Konstantin Sofiiuk \\
  Samsung AI \\
  \texttt{k.sofiiuk@samsung.com} \\
  \And
  Danila Rukhovich \\
  Samsung AI \\
  \texttt{d.rukhovich@samsung.com} \\
  \And
  Mikhail Romanov \\
  Samsung AI \\
  \texttt{m.romanov@samsung.com} \\
  \And
  Olga Barinova \\
  Samsung AI \\
  \texttt{o.barinova@samsung.com} \\
  \And
  Anton Konushin \\
  Samsung AI \\
  \texttt{a.konushin@samsung.com} \\
}

\begin{document}

\maketitle

\begin{abstract}
In recent years generative models of visual data have made a great progress, and now they are able to produce images of high quality and diversity. In this work we study representations learnt by a GAN generator. First, we show that these representations can be easily projected onto semantic segmentation map using a lightweight decoder. We find that such semantic projection can be learnt from just a few annotated images. Based on this finding, we propose \emph{LayerMatch} scheme for approximating the representation of a GAN generator that can be used for unsupervised domain-specific pretraining. We consider the semi-supervised learning scenario when a small amount of labeled data is available along with a large unlabeled dataset from the same domain. We find that the use of LayerMatch-pretrained backbone leads to superior accuracy compared to standard supervised pretraining on ImageNet. Moreover, this simple approach also outperforms recent semi-supervised semantic segmentation methods that use both labeled and unlabeled data during training. Source code for reproducing our experiments will be available at the time of publication.
\end{abstract}

\section{Introduction}

Generative models of visual data, and generative adversarial nets (GANs) in particular, have made remarkable progress in recent years \cite{goodfellow2014generative,arjovsky2017wasserstein,mescheder2018training,miyato2018spectral,karras2017progressive,brock2018large,kingma2018glow,karras2019style,karras2019analyzing}, and now they are able to produce images of high quality and diversity. Generative models have long been considered as a means of representation learning, with common assumption that the ability to generate data from some domain implies understanding of the semantics of that domain. Thus, various ideas about using GANs for representation learning have been studied in the literature \cite{radford2015unsupervised,chen2016infogan}. Most of these works are focused on producing universal feature representations by training a generative model on a large and diverse dataset \cite{donahue2016adversarial,donahue2019large}. However, the use of GANs as universal feature extractors has several limitations.

In this work we consider the task of unsupervised \emph{domain-specific pretraining}. Rather than trying to learn a universal representation on a diverse dataset, we focus on producing a specialized representation for a particular domain. Our intuition is that GAN generators are most efficient for learning \emph{high-resolution representations}, as generating a realistically-looking image implies learning appearance and location of different semantic parts. Thus, we experiment with semantic segmentation as a target downstream task. To illustrate our idea, we perform experiments with \emph{semantic projection} of GAN generator and show that it can be easily converted into a semantic segmentation model. Based on this finding, we introduce a novel \emph{LayerMatch} scheme that trains a model to predict the activations of the internal layers of GAN generator. Since the proposed scheme is trained on synthetic data and requires only a trained generator model, it can be used for unsupervised domain-specific pretraining.

As a practical use-case we consider the scenario when a limited amount of labeled data is available along with a large unlabeled dataset. This scenario is usually addressed by semi-supervised learning methods that use both labeled and unlabeled data during training. We evaluate LayerMatch pretraining as follows. First, a GAN model is trained on the unlabeled data, and a backbone model is pretrained using LayerMatch with the available GAN generator. Then, a semantic segmentation model with the LayerMatch-pretrained backbone is fine-tuned on the labeled part of the data. We perform experiments on two datasets with high quality GAN models available (CelebA-HQ \cite{lee2019maskgan} and LSUN \cite{yu2015lsun}). Surprisingly, we find that LayerMatch pretraining outperforms both the standard supervised pretraining on ImageNet and the recent semi-supervised semantic segmentation methods based on pseudo-labeling \cite{lee2013pseudo} and adversarial training \cite{hung2018adversarial}.

The rest of the paper is organized as follows. In Section \ref{sec:semantic_projection} we explore semantic projections of GAN generator and describe our experiments. In Section \ref{sec:layermatch} we introduce the LayerMatch scheme for unsupervised domain-specific pretraining that is based on inverting GAN generator. Section \ref{sec:layermatch_experiments} describes our experiments with the models pretrained using LayerMatch. In Section \ref{sec:related_work} we discuss related works.

\section{Semantic projection of a GAN generator}

Let us introduce the following notation. A typical GAN model consists of a jointly trained generator $G$ and a discriminator $D$. Generator $G$ transforms a random latent vector $l \in \mathbb{R}^k$ into an image $I_{gen} \in \mathbb{R}^{3 \times H \times W}$ and discriminator $D: \mathbb{R}^{3 \times H \times W} \rightarrow \mathbb{R}$ classifies whether an image is real or fake. Let us denote the activations of internal layers of $G$ for latent vector $l$ by $\mathbf{\Phi}(l)$.

\begin{figure}[t]
\setlength{\tabcolsep}{1pt}
\centering
  \begin{tabular}{cc}
         \includegraphics[width=0.45\linewidth]{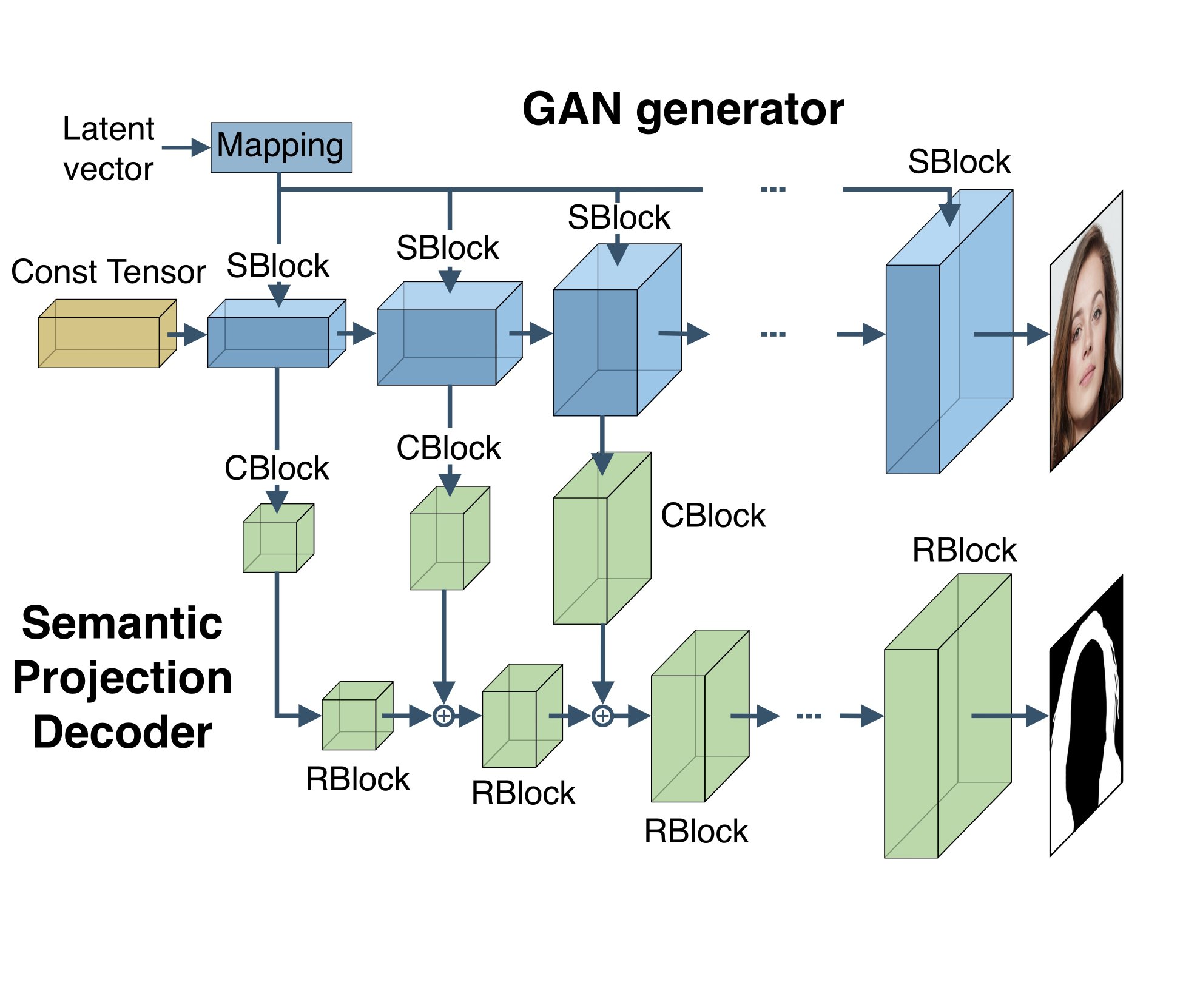} &
         \includegraphics[width=0.45\linewidth]{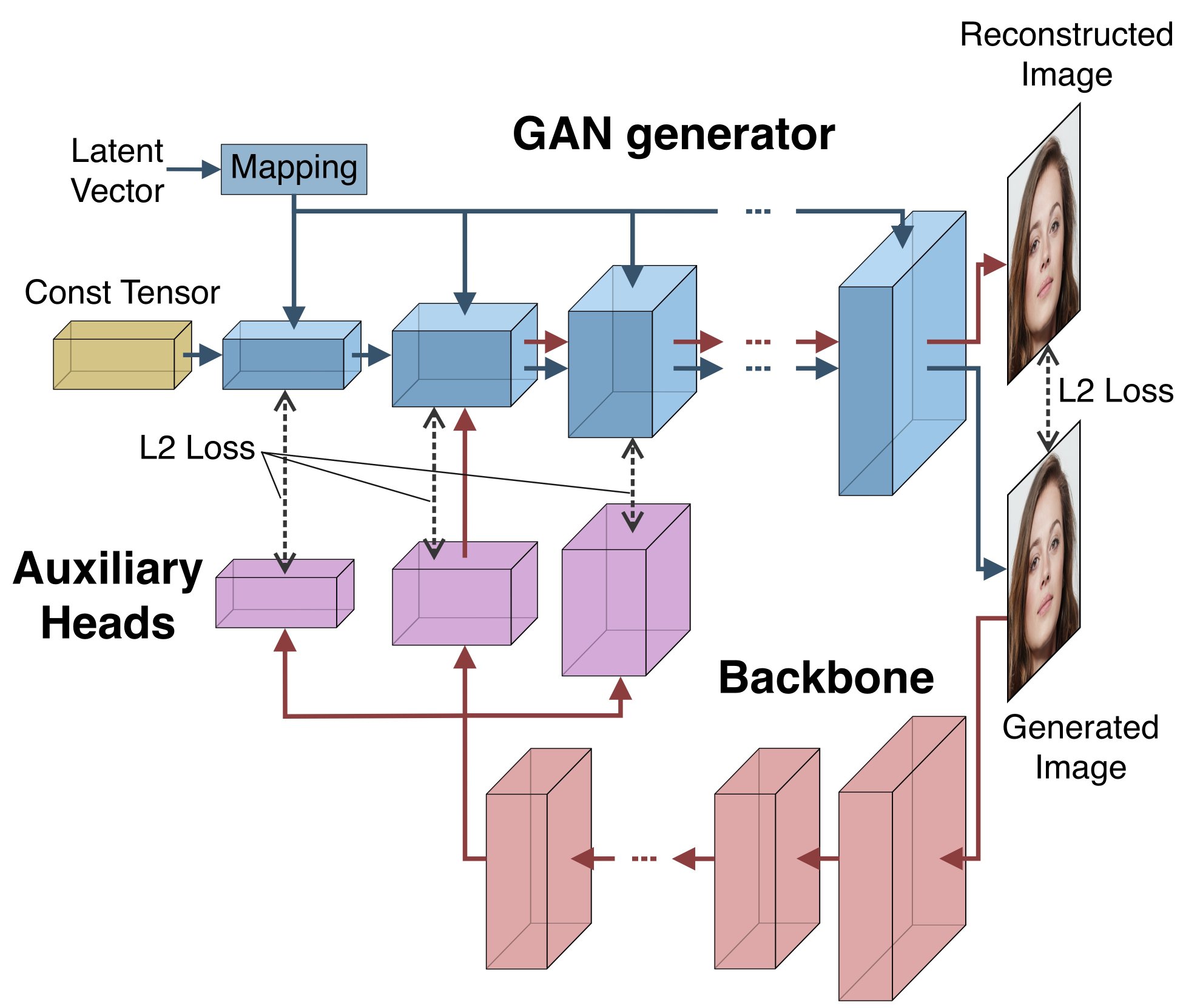} \\
         (a) & (b) \\
    \end{tabular}
    \caption{\small (a) Semantic projection implemented by a decoder built on top of a style-based generator as described in Section \ref{sec:semantic_projection}; (b) LayerMatch scheme for pretraining a backbone to approximate the activations of a GAN model as described in Section \ref{sec:layermatch}.}
    \label{fig:scheme_simple}
\end{figure}

\emph{Semantic projection} of a generator $P$ is a mapping of the features $\mathbf{\Phi}(l)$ onto the dense label map $L \in \{1,\dots, C\}^{W \times H}$ where $C$ is the number of classes. It can be implemented as a decoder that takes the features from different layers of a generator and outputs the semantic segmentation result. An example of a decoder architecture built on top of a style-based generator is shown in ~Figure~\ref{fig:scheme_simple} (a).

\subsection{Converting semantic projection model into a segmentation model}
\label{sec:semantic_projection}


Training procedure of a semantic projection model is shown in Algorithm~1. First, we sample a few images using GAN generator $G$ and store corresponding activations  $\{\mathbf{\Phi}_i\},  \mathbf{\Phi}_i = (\phi^i_1, \dots, \phi^i_k), i=1 \dots n$ of internal layers. The latent vectors are sampled from normal distribution. Then, we manually annotate a few generated images. The decoder is trained in a supervised manner using the segmentation masks from the previous step with corresponding intermediate generator features. We use cross-entropy between the predicted mask and ground truth as a loss function.

\begin{algorithm}[t]
    \DontPrintSemicolon
    \caption{Training semantic projection model}
    \KwInput{GAN model $(G, D)$}
    \KwOutput{Semantic projection model $P$}
    \Indp
    Generate $n$ images from the random latent vectors $I_i = G(l_i), l_i \sim \mathcal{N}(\mathbf{0}, \mathbf{\sigma})$, $i=1\dots n$ and store them along with their features $\{(I_i,\mathbf{\Phi}_i)\}$, $i=1\dots n$ \\
    Annotate the images and create semantic maps $L_i$, $i=1\dots n$ \\
    Train a decoder $P$ on pairs $\{(\mathbf{\Phi}_i,L_i)\}$, $i=1\dots n$ \\
    \label{alg:projection}
\end{algorithm}

Once we have trained a semantic projection model $P$, we can obtain pixelwise annotation for generated images. For this purpose, we can apply $P$ to the features produced by generator $L_{gen}=P(\mathbf{\Phi})$. However, the features of a generator are not available for real images. Since semantic projection alone does not allow obtaining semantic segmentation maps for real images, we propose Algorithm~2 for converting the semantic projection model into a semantic segmentation model applicable to real images. The intuition is that training on a large number of GAN-generated images along with accurate annotations provided by semantic projection should result in an accurate segmentation model.

\begin{algorithm}
\caption{Converting semantic projection into semantic segmentation model}
    \KwInput{GAN model $(G, D)$, semantic projection model $P$}
    \KwOutput{Semantic segmentation model $S$}
    \Indp Generate $N$ images from the random latent vectors $I_i=G(l_i), l_i \sim \mathcal{N}(\mathbf{0}, \mathbf{\sigma})$ $i=1\dots N$ and store them along with their features $\{(I_i,\mathbf{\Phi}_i)\}$, $i=1\dots N$ \\
    Compute results of semantic projection $\{P(\mathbf{\Phi}_i)\}$, $i=1\dots N$ \\
    Train semantic segmentation model $S$ on pairs $\{(\mathbf{\Phi}_i,P(\mathbf{\Phi}_i))\}$, $i=1\dots N$ \\
    \label{alg:segmentation}
\end{algorithm}

\subsection{Experiments with semantic projections}
\label{sec:experiments_semantic_projection}

 In this section we address the following questions: 1) Will a lightweight decoder be sufficient to implement an accurate semantic projection model? 2) How many images are required to train semantic projection to a reasonable accuracy? 3) Will the use of Algorithm~2 lead to improved performance on real images? 

\textbf{Experimental protocol.} We perform experiments with style-based generator \cite{karras2018style} on two datasets (FFHQ and LSUN-cars). In both experiments, we manually annotate 20 randomly generated images for training the semantic projection models. For FFHQ experiment we use two classes: hair and background. Hair is a challenging category for segmentation as it usually has difficult shape with narrow elongated parts. For LSUN-cars we use car and background categories. We also train DeepLabV3+ \cite{deeplabv3plus2018} model using Algorithm~2 with semantic projection models trained on 20 images. In all experiments we use ResNet-50 as a backbone. For LSUN-cars we experiment with both ImageNet-pretrained and randomly initialized backbones. For comparison we train a similar DeepLabV3+ model on 20 labeled real images. 80 annotated real images are used for testing the semantic segmentation models. Pixel accuracy and intersection-over-union (IoU) are measured for methods comparison. 

\textbf{Architecture of the semantic projection model.} The lightweight decoder architecture for semantic projection is shown in ~Figure~\ref{fig:scheme_simple}. It has an order of magnitude fewer parameters compared to the standard decoder architectures and 16 times fewer than DeepLabV3+ decoder. Each CBlock of the decoder takes the features from corresponding SBlock of StyleGAN as an input. CBlock consists of a 50\% dropout, a convolutional and a batch normalization layers. Each RBlock of a decoder has one residual block with two convolutional layers. The number of feature maps in each convolutional layer of the decoder is set to 32, as wider feature maps resulted in just minor improvement in our experiments.


\begin{figure*}[t]
\setlength{\tabcolsep}{1pt}
\centering
  \begin{tabular}{cc}
      \includegraphics[height=0.2\linewidth]{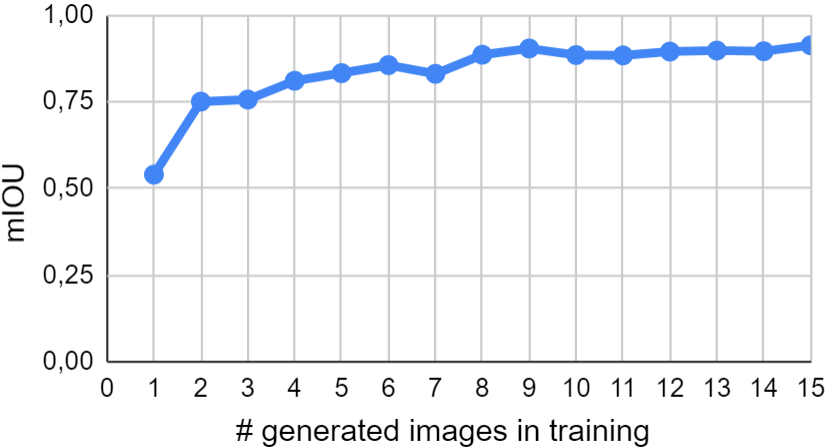} &
      \includegraphics[height=0.2\linewidth]{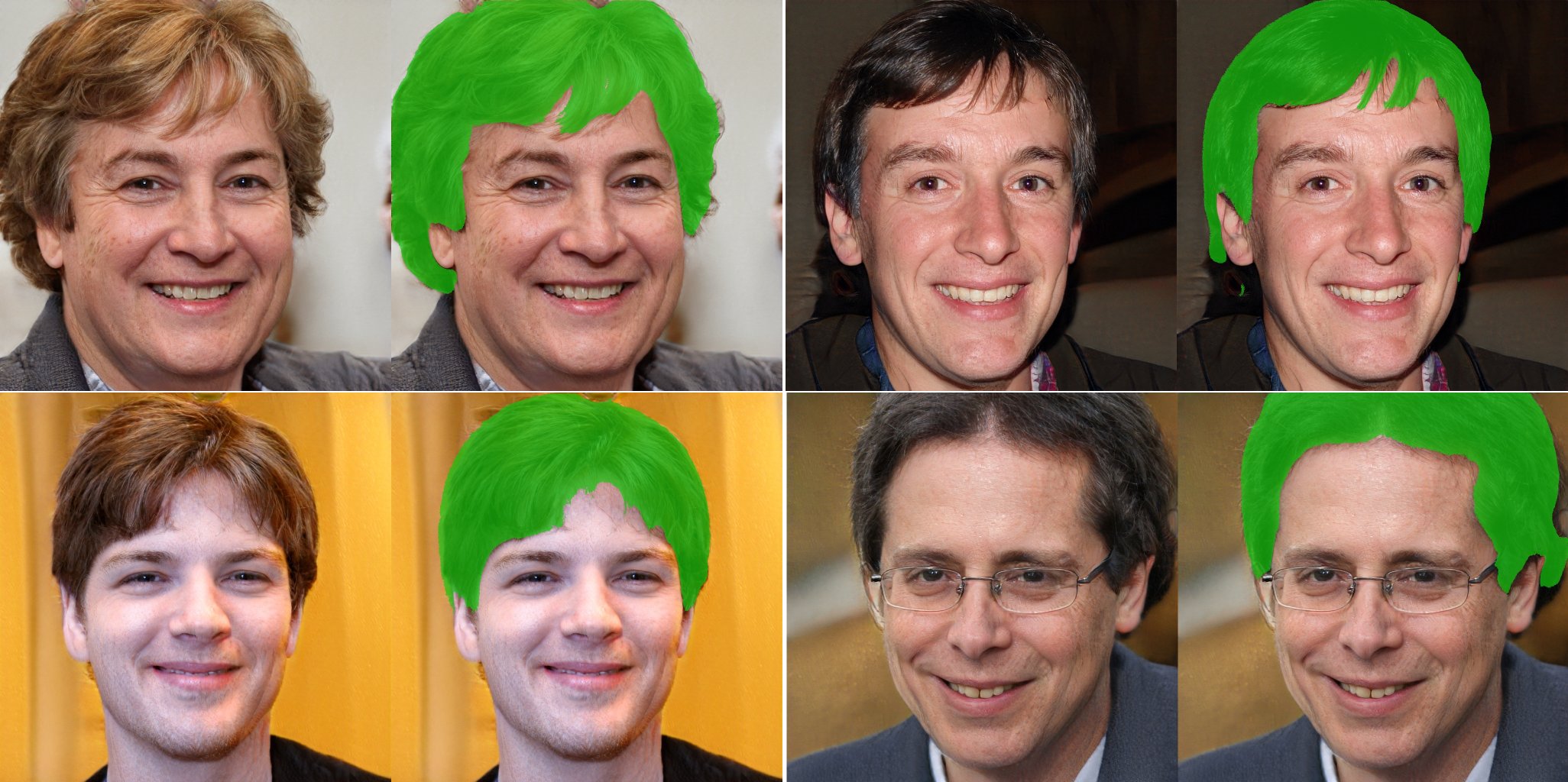}\\
      (a) & (b) \\
  \end{tabular}
  \caption{\small (a) - evaluation results of the semantic projection model on two classes (background and hair) with respect to the number of images in training; (b) - outputs of semantic projection model for test images generated by StyleGAN. Note that while the model was trained just on 20 images, it provides quite accurate segmentation.}
  \label{fig:projection_examples_ffhq}
  
\hspace{12pt}

\setlength{\tabcolsep}{1pt}
\centering
  \begin{tabular}{ccc}
      \includegraphics[height=0.1\linewidth]{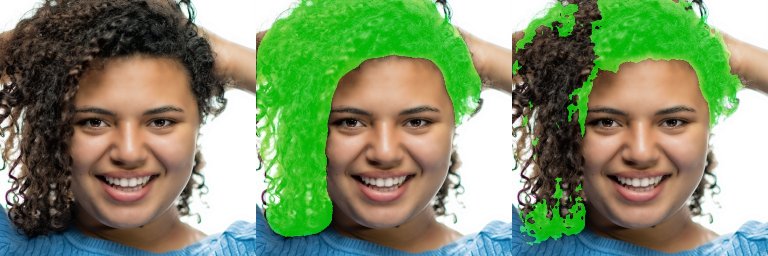} &
      \includegraphics[height=0.1\linewidth]{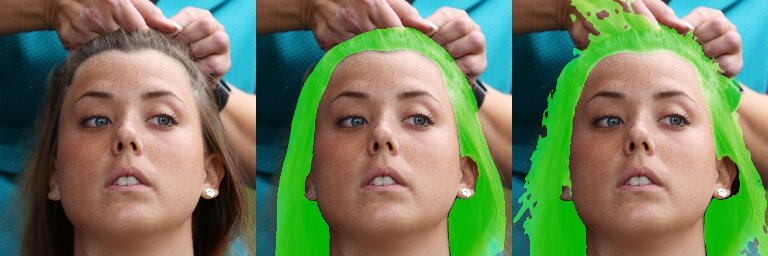} &
      \includegraphics[height=0.1\linewidth]{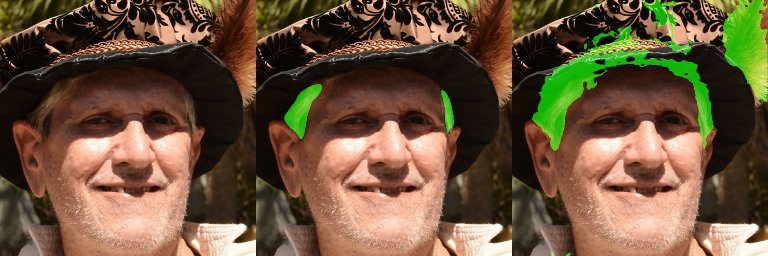} \\
  \end{tabular}
  \begin{tabular}{cc}
    \includegraphics[height=0.102\linewidth]{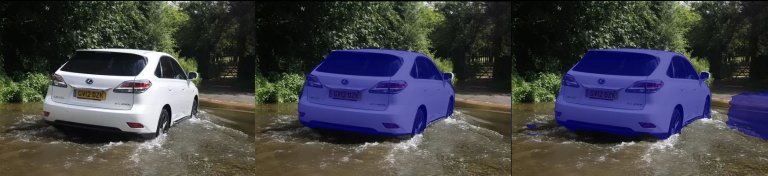} &
    \includegraphics[height=0.102\linewidth]{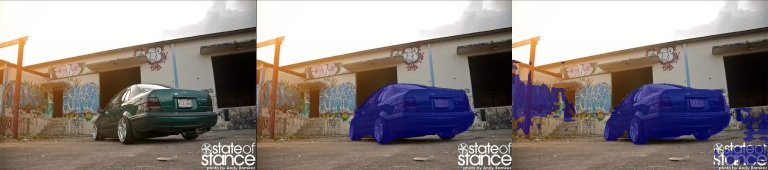} \\
  \end{tabular}
  \caption{\small Outputs of two semantic segmentation models trained with equal amount of supervision. First row - FFHQ, second row - LSUN-cars. From left to right: input image, output of the model trained with Algorithm~2 on synthetic images, output of the model trained on the same number of real images.}
  \label{fig:segmentation_examples}
\end{figure*}

\begin{table*}[t]
    \begin{center}
    \small
    \begin{tabular}{c|l|c|c|c}
        \hline
        Categories & Method & \makecell{ImageNet-pretrained \\ backbone} & accuracy & IoU \\
        \hline
        \hline
        \multirow{2}{*}{Hair/Background} & Training on 20 labeled images & \multirow{2}{*}{+} & 0.9515 & 0.8194 \\
         & Algorithm~2 with 20 labeled images&  & \textbf{0.9675} & \textbf{0.8759} \\
        \hline
        \hline
        \multirow{4}{*}{Car/Background} & Training on 20 labeled images & \multirow{2}{*}{-} & 0.8588 & 0.6983 \\
        & Algorithm~2 with 20 labeled images &  & 0.9787 & 0.9408 \\
        \cline{2-5}
        & Training on 20 labeled images & \multirow{2}{*}{+} & 0.9641 & 0.9049 \\
        & Algorithm~2 with 20 labeled images &  & \textbf{0.9862} & \textbf{0.9609} \\
        \hline
    \end{tabular}
    \end{center}
    \caption{\small Comparison of the segmentation models trained with equal amount of supervision. See text for more details.}
    \label{tab:evaluation_hair_car}
\end{table*}

\textbf{Results and discussion.} Figure~\ref{fig:projection_examples_ffhq} (b) shows outputs of a semantic projection model trained on 20 synthetic images using Algorithm~1. The results of varying the size of a training set from 1 to 15 synthetic images is shown in Figure~\ref{fig:projection_examples_ffhq} (a). The test set in this experiment contains 30 manually annotated GAN-generated images. We observe that even with a single image in training set, the model achieves reasonable segmentation quality, and the quality grows quite slowly after 10 images.

Next, we compare two semantic segmentation models trained with equal amount of supervision. The first one uses Algorithm~2 with semantic projection model trained on 20 synthetic images. The second one uses ImageNet-pretrained backbone and is trained on 20 real images. Table~\ref{tab:evaluation_hair_car} shows quantitative comparison of the two models. One can notice that in case when the backbone for DeepLabV3+ is randomly initialized, the model trained with Algorithm~2 is significantly more accurate compared to the baseline approach. When using ImageNet-pretrained backbones, Algorithm~2 leads to 6\% improvement in terms of IoU for both datasets. Figure \ref{fig:segmentation_examples} shows examples of hair and car segmentation for real images from the test set.

Our experiments of two datasets demontrate that a lightweight decoder is sufficient to implement an accurate semantic projection model. We observe that just a few annotated images are enough to train semantic projection to a reasonable accuracy. The Algorithm~2 leads to improved accuracy on real images compared to simply training a similar model with the same number of annotated images.

\section{Transfer learning using generator representation}

Training a semantic projection model introduced in Section \ref{sec:semantic_projection} requires manual annotation of GAN-generated images. Thus, we cannot use standard real-image datasets for comparison with other works. Real images could potentially be embedded into the GAN latent space, but in practice this approach has its own limitations \cite{bau2019seeing}. Besides, some of the images produced by GAN generators can be hard to label. 

A semantic segmentation network transforms an image $I \in \mathbb{R}^{3 \times H \times W}$ into a segmentation map. At the same time a GAN generator $G$ transforms a random vector $l \in \mathbb{R}^k$ into an image $I_{gen} \in \mathbb{R}^{3 \times H \times W}$. Obviously, the input dimensions of these two types of models do not match. Therefore, the models trained for image generation cannot be directly applied to image segmentation. To overcome this issue one can think of inverting a generator. Inverted GAN generators have been widely used for the task of image manipulation \cite{abdal2019image2stylegan,bau2019seeing}. For this purpose, an encoder model is usually trained to predict the latent vector from an image. Following \cite{abdal2019image2stylegan,bau2019seeing} we train an encoder network, but predict the activations of a fixed GAN generator instead of the latent vector. The backbone of the trained encoder can then be used to initialize a semantic segmentation model. 

\subsection{Unsupervised pretraining with LayerMatch}
\label{sec:layermatch}

The scheme of the LayerMatch algorithm is shown in~Figure~\ref{fig:scheme_simple} (b). 
We can view generator $G$ as a function of the latent vector $l$ and all the intermediate activations: $G = G(l, \phi_1, \phi_2, .., \phi_n)$, where intermediate features themselves depend on the latent vector and all the previous features: $\phi_i = \phi_i(l, \phi_1, \phi_2, .., \phi_{i-1})$. The generated image $I_{gen}$ is fed to the encoder $E$, which tries to predict the $n$ specified activation tensors: $\hat{\Phi} = E(I_{gen})$, where $\hat{\Phi} = (\hat{\phi}_1, \hat{\phi}_2,\dots,\hat{\phi}_n)$.

\begingroup

The loss function for LayerMatch training consists of two terms: 
\begin{equation}
    \mathcal{L} = \mathcal{L}_{rec} + \mathcal{L}_{match},
    \nonumber
\end{equation}

where matching loss $\mathcal{L}_{match}$ is the sum of the L2-losses between generated and predicted features that penalizes difference between the outputs of the encoder and the activations of the generator:

\begin{equation}
    \mathcal{L}_{match} = \frac{1}{n} \sum_{i=1}^n {\|\phi_i - \hat{\phi}_i\|}_2^2
    \nonumber
\end{equation}

Reconstructed image $I_{rec}$ is obtained by replacing random feature $\phi_m$ with $\hat{\phi}_m$, where $1 \le m \le n$, and recalculating features $\tilde{\phi_j}=\phi_j(l, \phi_1, \dots, \phi_{m-1}, \hat{\phi}_m,\tilde{\phi}_{m+1} \dots \tilde{\phi_{j-1}}) , m < j \le n$: 

\begin{equation}
    I_{rec} = G(l, \phi_1, \dots , \phi_{m-1}, \hat{\phi}_m, \tilde{\phi}_{m+1}, .., \tilde{\phi}_n)
    \nonumber
\end{equation}

Reconstruction loss $\mathcal{L}_{rec}$ is the L2-loss between the generated image and the reconstructed image. This loss controls that the generator produces an image which is close to the original one when generator activations are replaced with the outputs of the backbone:
\begin{equation}
    \mathcal{L}_{rec} = {\|I_{rec} - I_{gen}\|}_2^2
    \nonumber
\end{equation}

\endgroup



\begin{figure}[t] 
\setlength{\tabcolsep}{3pt}
\centering
    \begin{tabular}{ccc}
         \includegraphics[height=0.25\linewidth]{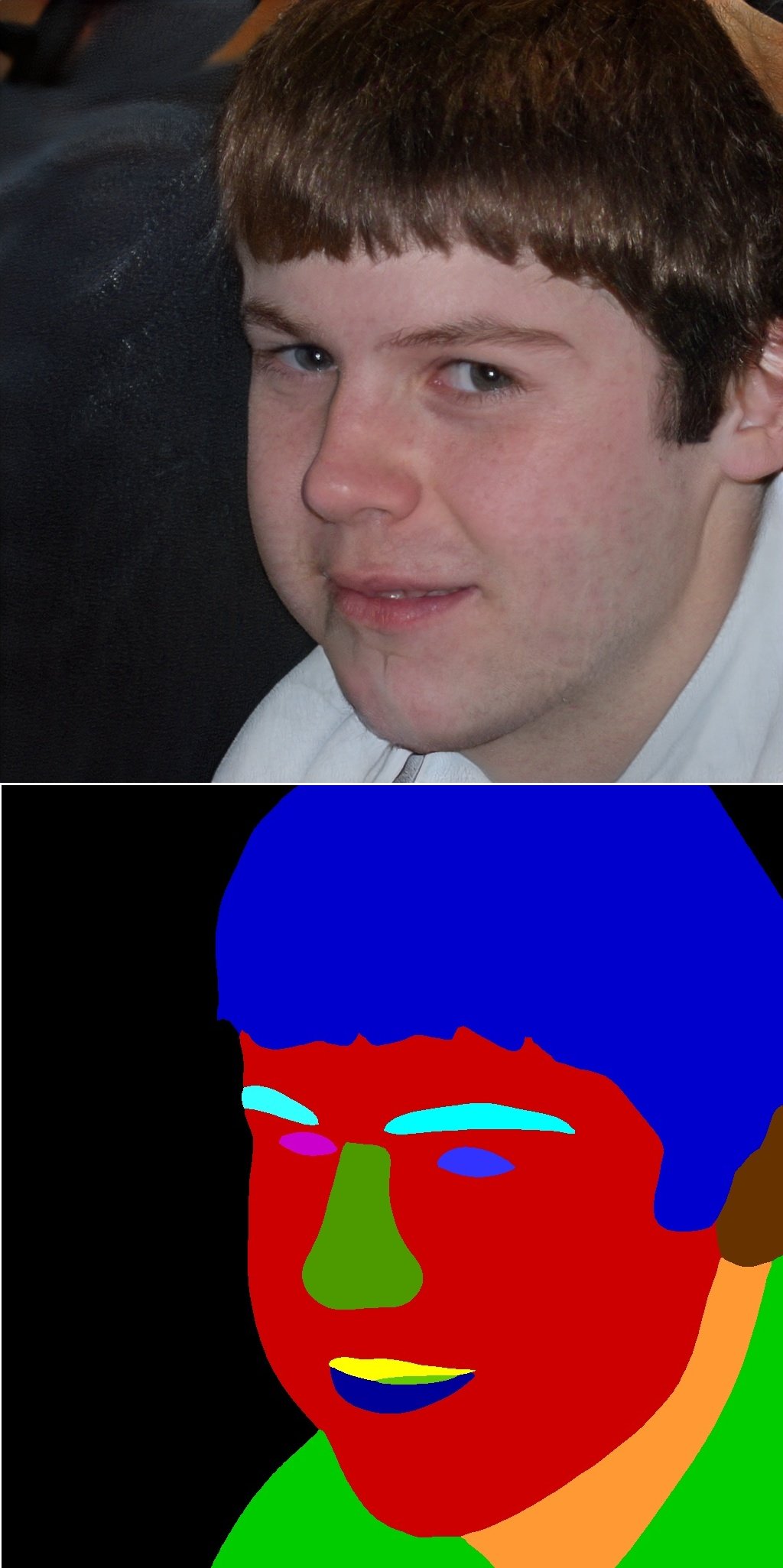} &
         \includegraphics[height=0.25\linewidth]{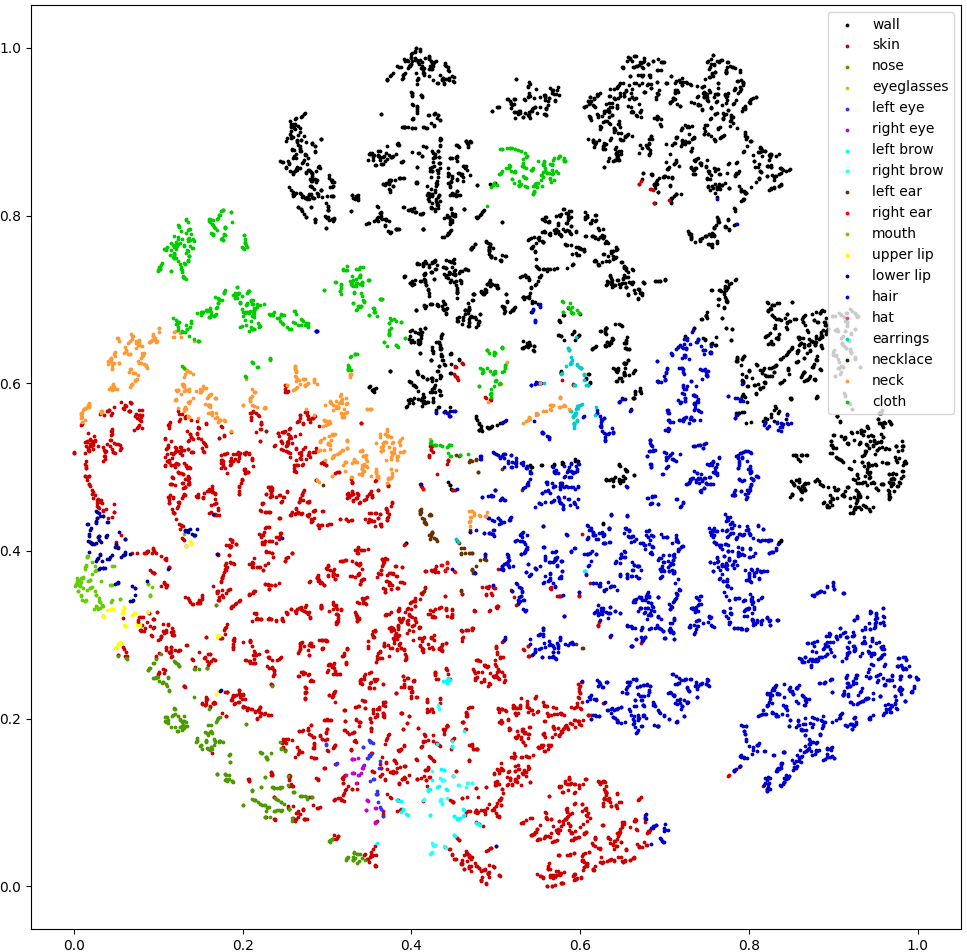} &
         \includegraphics[height=0.25\linewidth]{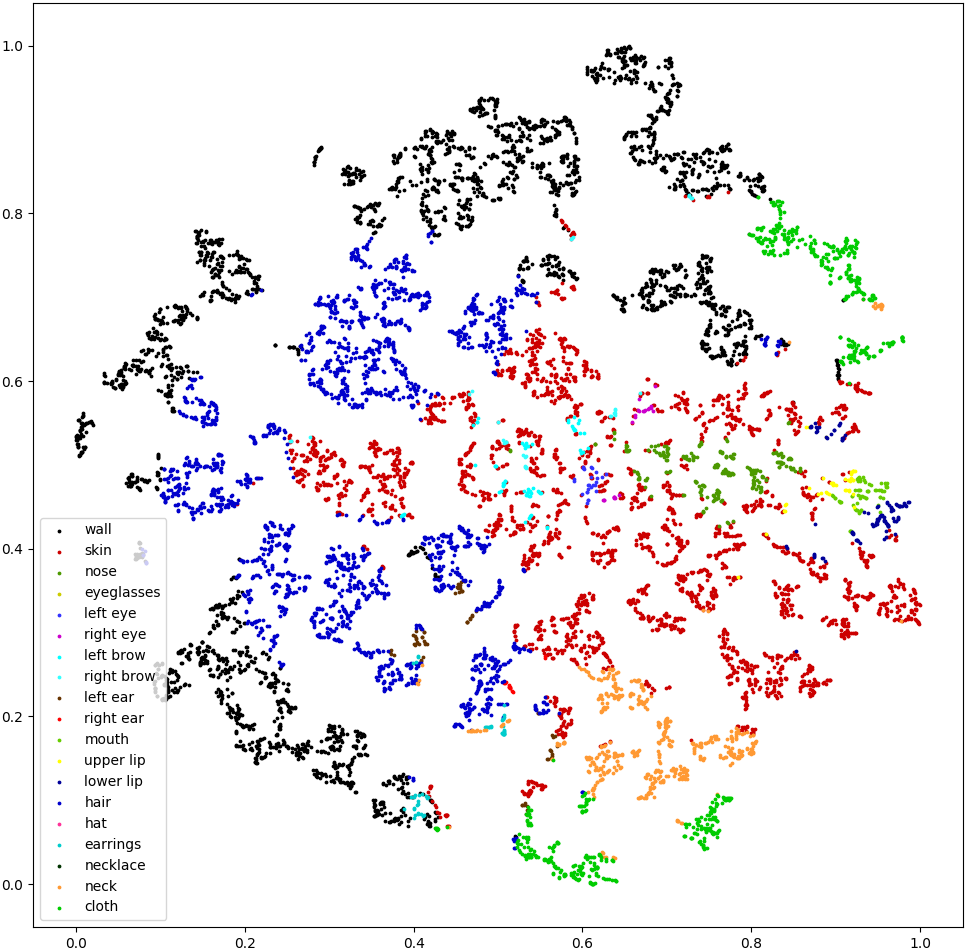} \\
         (a) & (b) & (c)
    \end{tabular}
    \caption{\small (a) - example face image with color-coded semantic annotation, (b) - t-SNE visualization of the features of an ImageNet-pretrained backbone, (c) - similar visualization of the features learnt with LayerMatch. In (b) and (c) each point is color-coded according to the ground truth annotation. See text for more details.}
    \label{fig:tsne_gan}
\end{figure}

Figure~\ref{fig:tsne_gan} shows t-SNE visualizations of the features from the internal layers of two similar models. Each point on the plots (b) and (c) represents a feature vector corresponding to a particular position in an image. We used the activations for 5 images in this visualization. Plot (a) shows the activations of an ImageNet-pretrained model, and the plot (b) shows activations of the model pretrained with LayerMatch. One can observe that the distribution of the features learned by LayerMatch contains semantically meaningful clusters, while the universal features of an ImageNet-pretrained backbone look more scattered.

\section{Experiments with LayerMatch}
\label{sec:layermatch_experiments}

\textbf{Evaluation protocol.} The standard protocol for evaluation of unsupervised learning techniques proposed in \cite{zhang2016colorful} involves training a model on unlabeled ImageNet, freezing its learned representation, and then training a linear classifier on its outputs using all of the training set labels. This protocol is based on the assumption that the resulting representation is universal and applicable to different domains. We rather focus on domain-specific pretraining, or "specializing" the backbone to a particular domain. We aim at high-resolution tasks, \eg semantic segmentation. Therefore, we apply a different evaluation protocol.

We assume that we have a high-quality GAN model trained on a large unlabeled dataset from the domain of interest along with a limited number of annotated images from the same domain. The unlabeled data is used for training a GAN model, which in turn is used for pretraining the backbone model using LayerMatch (see Algorithm~3). The pixelwise-annotated data is later used for training a fixed semantic segmentation network with the pretrained backbone using a standard cross-entropy loss. Then, we evaluate the resulting model on a test set across the standard semantic segmentation metrics such as mIOU and pixel accuracy. We perform experiments with varying fraction of labeled data. In all our experiments we initialize the networks with ImageNet-pretrained backbones.

\begin{algorithm}
    \DontPrintSemicolon
    \caption{Training semantic projection model}
    \KwInput{GAN model $(G, D)$ trained on a large unlabeled dataset, a small labeled dataset ${(I_j, L_j)}, j=1\dots n$}
    \KwOutput{Semantic segmentation model $S$}
    \Indp
    Generate $N$ images from the random latent vectors $I_i=G(l_i), l_i \sim \mathcal{N}(\mathbf{0}, \mathbf{\sigma})$, $i=1\dots N$ and store them along with their features $\{(I_i,\mathbf{\Phi}_i)\}$, $i=1\dots N$ \\
    Train the backbone using LayerMatch using the pairs $\{(I_i,\mathbf{\Phi}_i)\}$, $i=1\dots N$ \\
    Train a semantic segmentation model $S$ on the labeled part of the data $\{(I_j,L_j)\}$, $j=1\dots n$ \\
    \label{alg:layermatch_segmentation}
\end{algorithm}

\textbf{Comparison with prior work.}
For all compared methods we use the same network architectures differing only in training procedure and loss functions used. The first baseline uses a standard ImageNet-pretrained backbone without domain-specific pretraining. The semantic segmentation model is trained using available annotated data and does not use the unlabeled data. 

The other two baselines are recent semi-supervised segmentation methods using both labeled and unlabeled data during training. In the experiments with these methods we used exactly the same amount of both labeled and unlabeled data as for LayerMatch. Namely, for the experiments with Celeba-HQ we used both the unlabeled part of CelebA and the FFHQ dataset, that was used for GAN training. For the experiments with LSUN-church all the unlabeled data in LSUN-church dataset was used during training.

The first semi-supervised segmentation method that we use for comparison is based on pseudo-labeling \cite{lee2013pseudo}. Unlabeled data is augmented by generating pseudo-labels using the network predictions. Only the pixels with high-confidence pseudo-labels are used as ground truth for training. The second one is an adversarial semi-supervised segmentation approach \cite{hung2018adversarial}. In this method the segmentation network is supervised by both the standard cross-entropy loss with the ground truth label map and the adversarial loss with the discriminator network. In our experiments we used the official implementation provided by the authors, and changed only the backbone.

\begin{figure*}[t]
\centering
  \begin{tabular}{cc}
    \includegraphics[width=0.4\linewidth]{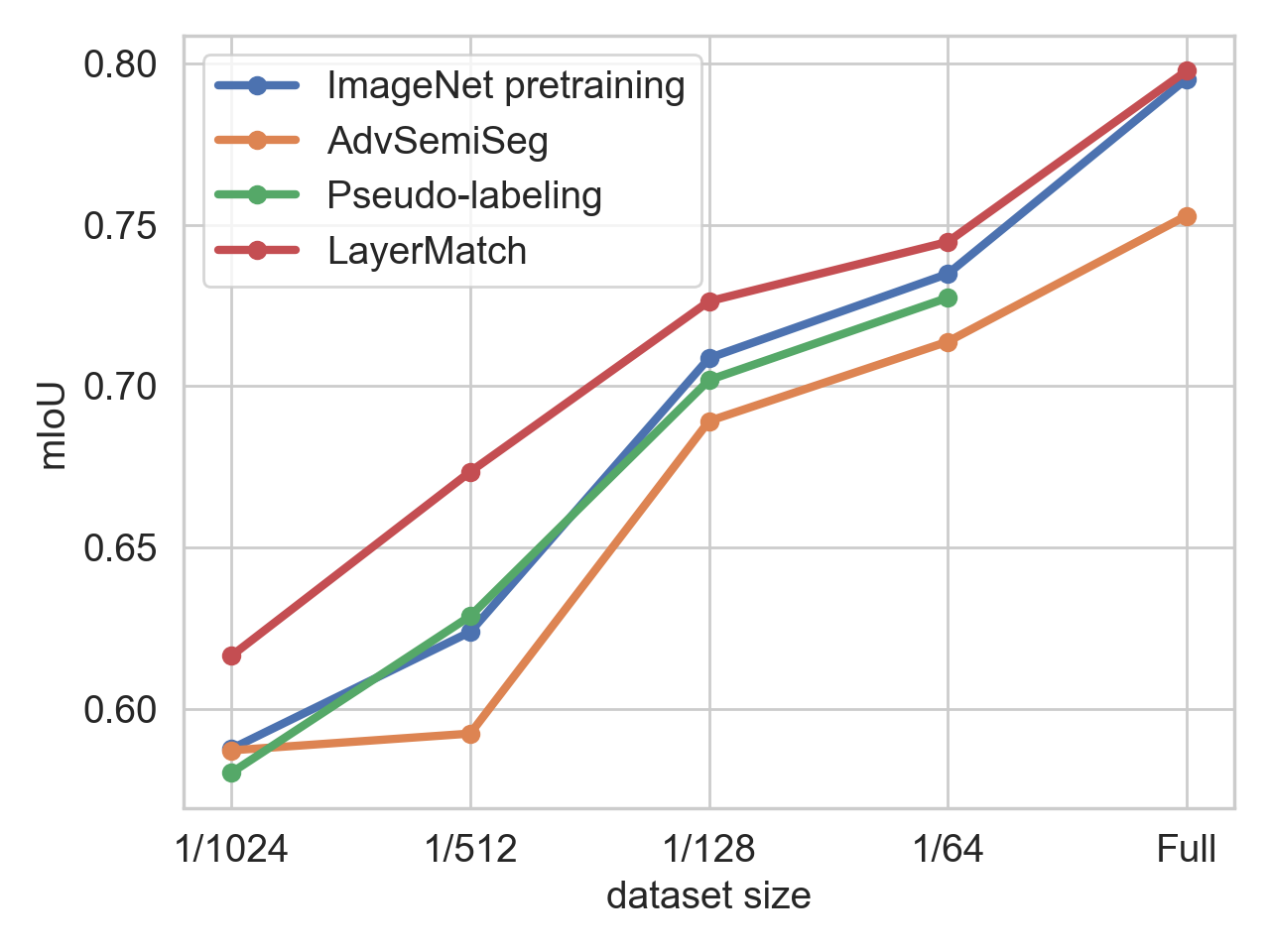} &
    \includegraphics[width=0.4\linewidth]{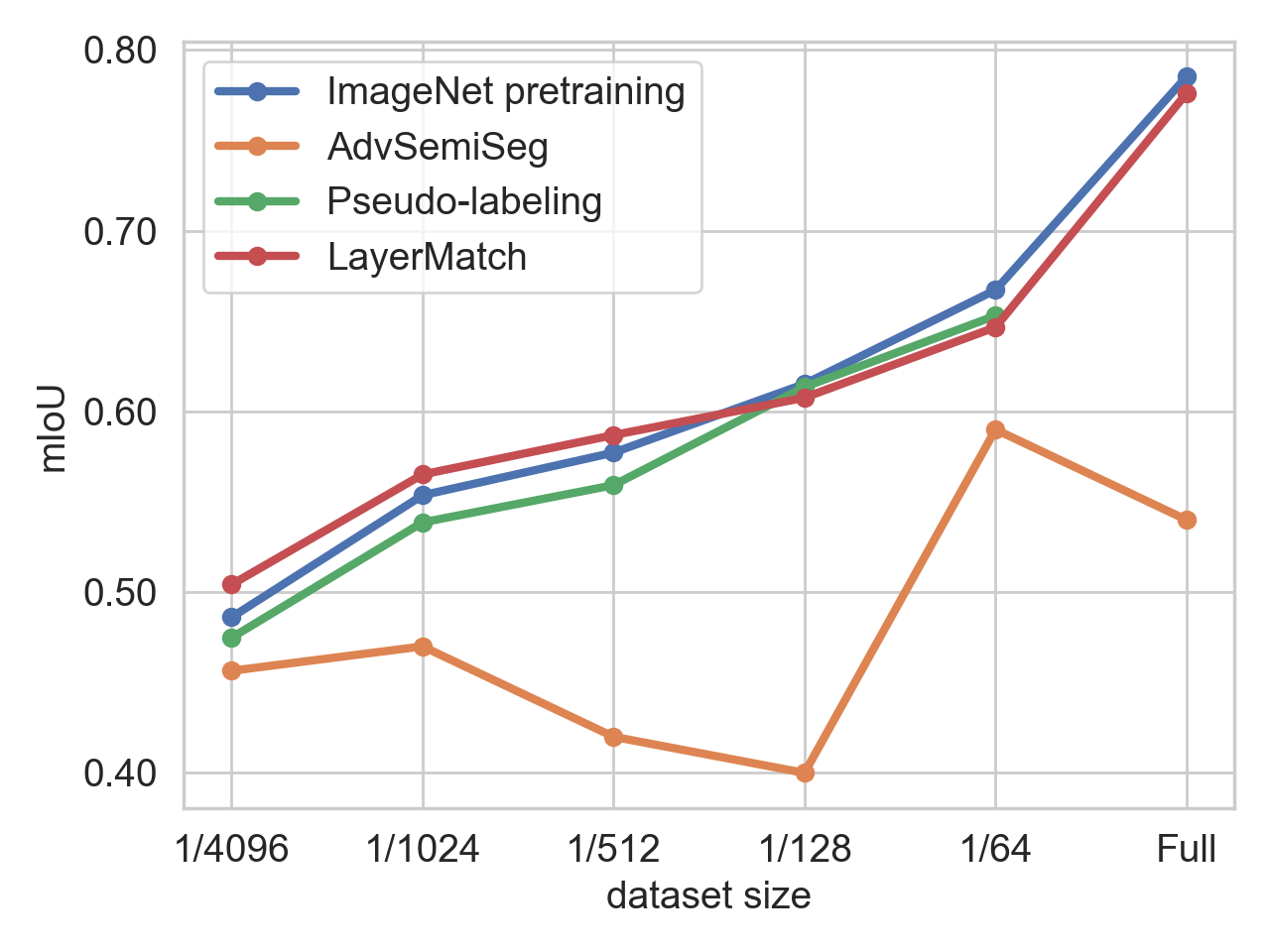} \\
    (a) & (b) \\
  \end{tabular}
  \caption{\small Comparison of the models trained with Algorithm~3 to semi-supervised segmentation methods for a varying number of annotated samples. (a) - FFHQ+CelebA-HQ dataset. (b) - LSUN-church dataset}
  \label{fig:figure_results}
\end{figure*}

\textbf{Datasets.}
Celeba-HQ \cite{lee2019maskgan} contains 30,000 high-resolution face images selected from the CelebA dataset \cite{liu2015faceattributes}, each image having a segmentation mask with the resolution of 512x512 and 19 classes including all facial components and accessories such as skin, nose, eyes, eyebrows, ears, mouth, lips, hair, hat, eyeglass, earring, necklace, neck, cloth, and background. We use a StyleGAN2 model trained on FFHQ dataset provided in \cite{karras2019analyzing} that has a FID measure 3.31 and PPL 125. In the experiments with Celeba-HQ we vary the fraction of labeled data from 1/1024 to the full dataset. 

LSUN-church \cite{yu15lsun} contains 126,000 images of churches of 256x256 resolution. We have selected top 10 semantic categories that occupy more than 1\% of image area, namely road, vegetation, building, sidewalk, car, sky, terrain, pole, fence, wall. We use a StyleGAN2 model provided in \cite{karras2019analyzing} that has a FID measure 3.86 and PPL 342. As LSUN dataset does not contain pixelwise annotation, we take the outputs of the Unified Scene Parsing Network \cite{xiao2018unified} as ground truth in this experiment similarly to \cite{bau2019seeing}. In the experiments with Celeba-HQ we vary the fraction of labeled data from 1/4096 to the full dataset. 

\textbf{Implementation details.}
HRNet \cite{sun2019high} is used as an encoder architecture. We add $K$ auxiliary heads for each of $K$ activations that we want to predict (see Figure \ref{fig:scheme_simple} (b)). After training, auxiliary heads are discarded and only the pretrained backbone is used for transfer learning, similar to ImageNet pretraining. For pretraining the encoder we use Adam optimizer with the learning rate $10^{-4}$ and the cosine learning rate decay. We use source code from HRNet repository for training semantic segmentation networks.

\textbf{Results and discussion.} Figure \ref{fig:figure_results} shows the comparison of the proposed LayerMatch pretraining scheme to 3 baseline methods across 2 datasets with varying fraction of annotated data. Pseudo-labeling is applicable in case when some part of the dataset is unlabelled. 

\begin{figure*}[t]
\setlength{\tabcolsep}{1pt}
\centering
  \begin{tabular}{cc}
      \includegraphics[height=0.1\linewidth]{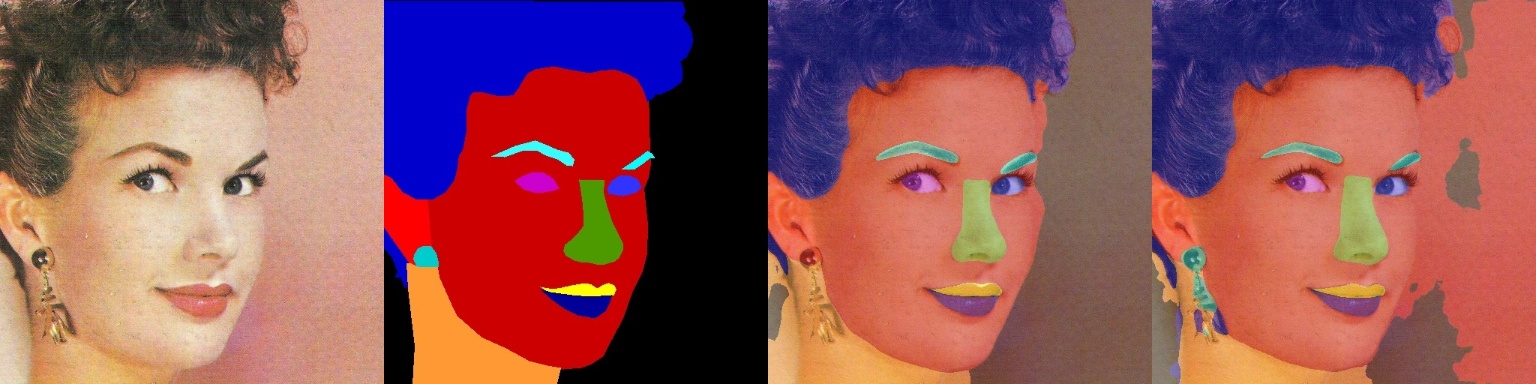} &
      \includegraphics[height=0.1\linewidth]{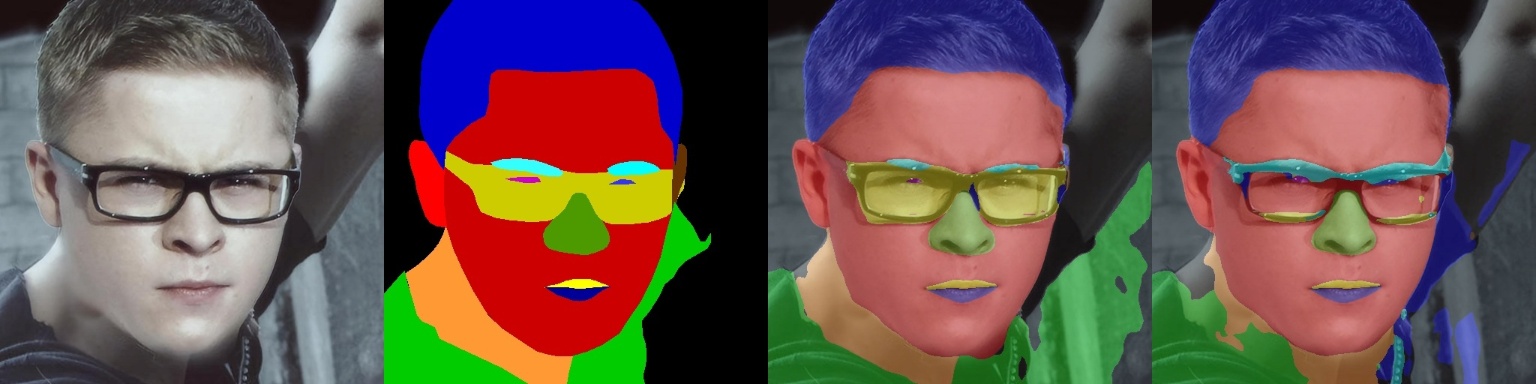} \
  \end{tabular}
  \begin{tabular}{cc}
      \includegraphics[height=0.1\linewidth]{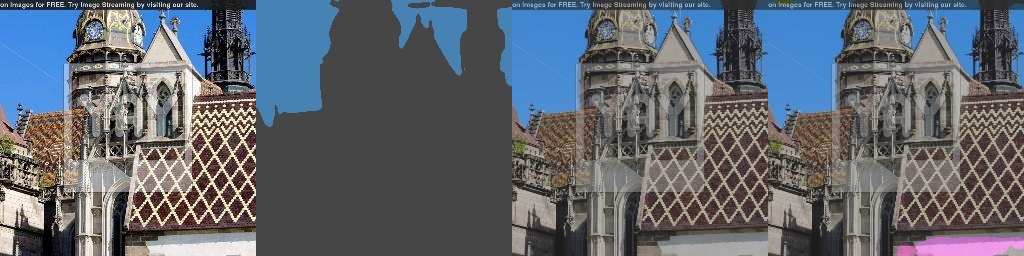} &
      \includegraphics[height=0.1\linewidth]{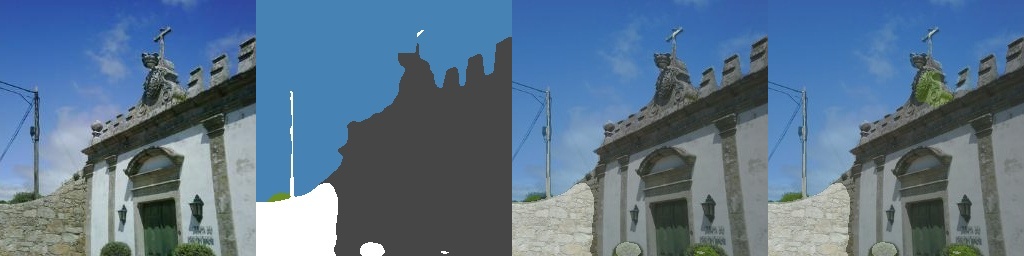} \\
  \end{tabular}
  \caption{\small Test results. First row: CelebA-HQ, 1/512, second row: LSUN-church, 1/512. From left to right: input image, LayerMatch pretraining, ImageNet pretraining only. }
  \label{fig:layermatch_examples}
\end{figure*}

One can see that LayerMatch pretraining shows significantly higher IoU compared to the baseline methods on Celeba-HQ (see Figure \ref{fig:figure_results} (a)) for any fraction of the labeled data. For LSUN-church it shows higher accuracy compared to other methods in cases when up to 1/512 of the data is annotated. Figure \ref{fig:layermatch_examples} shows qualitative comparison of the model pretrained with LayerMatch to the standard ImageNet pretraining on Celeba-HQ when trained with 1/512 of annotated data. The difference between two models is quite noticeable for both CelebA-HQ and for LSUN-church. Table \ref{tab:evaluation_layermatch} shows category-wise results for all four compared models trained with 1/512 of labeled data. LayerMatch pretraining leads to significant accuracy improvement for the eyeglasses category.

Overall, LayerMatch pretraining leads to improved results in semi-supervised learning scenario compared to both simple ImageNet pretraining and to semi-supervised segmentation methods. Lower accuracy for larger fraction of annotated datasets on LSUN-church can be attributed to lower quality of LSUN-church GAN generator compared to Celeba-HQ GAN generator. Another possible reason for this effect may be the imperfect annotation of both training and test data, which may lead to inaccuracies in evaluation.

\begin{table*}[t]
    \begin{center}
    \small
    \tabcolsep=0.08cm
    \begin{tabular}{c|cc|ccccccccccccccccccc}
       Method & \rotatebox{90}{pixAcc} & \rotatebox{90}{mIoU} & \rotatebox{90}{background} & \rotatebox{90}{skin} & \rotatebox{90}{nose} & \rotatebox{90}{eye glasses} & \rotatebox{90}{left eye} & \rotatebox{90}{right eye} & \rotatebox{90}{left brow} & \rotatebox{90}{right brow} & \rotatebox{90}{left ear} & \rotatebox{90}{right ear} & \rotatebox{90}{mouth} & \rotatebox{90}{upper lip} & \rotatebox{90}{lower lip} & \rotatebox{90}{hair} & \rotatebox{90}{hat} & \rotatebox{90}{earrings} & \rotatebox{90}{necklace} & \rotatebox{90}{neck} & \rotatebox{90}{cloth} \\
        \hline
        \hline
       ImageNet only & .92 &	.62 & .89 & .89 & \textbf{.87} & \underline{.01} & .75 & .77 & .67 & .68 & .71 & .72 & .76 & .74 & .79 & \textbf{.87} & 0. & \textbf{.34} & 0. & .77 & .62 \\
       AdvSemiSeg \cite{hung2018adversarial} & .91 & .62 & .89 & .89 & .85 & \underline{.14} & .75 & .77 & .66 & .66 & .71 & .70 & .77 & .73 & .78 & .86 & 0. & .29 & 0. & .75 & .59 \\
       Pseudo-labeling \cite{lee2013pseudo} & .92 & .63 & .89 & .90 & .86 & \underline{.01} & .77 & \textbf{.78} & \textbf{.71} & \textbf{.70} & .73 & .72 & \textbf{.79} & .75 & .79 & .86 & 0. & .33 & 0. & .77 & .58 \\
       \hline
       LayerMatch & \textbf{.93} & \textbf{.67} & \textbf{.90} & \textbf{.91} & \textbf{.87} & \underline{\textbf{.68}} & \textbf{.78} & \textbf{.78} & .70 & .69 & \textbf{.75} & \textbf{.74} & \textbf{.79} & \textbf{.76} & \textbf{.80} & \textbf{.87} & 0. & \textbf{.34} & 0. & \textbf{.80} & \textbf{.64} \\
    \end{tabular}
    \end{center}
    \caption{
    \small Comparison of segmentation models trained on CelebA-HQ dataset with equal amount of supervision. Notice that LayerMatch provides better results for almost all categories and improves the IoU for eye glasses category by several times. }
    \label{tab:evaluation_layermatch}
\end{table*}

\section{Related work}
\label{sec:related_work}


Several works consider generative models for unsupervised pretraining \cite{makhzani2015adversarial, larsen2015autoencoding, donahue2016adversarial, donahue2019large}. One of the approaches \cite{radford2015unsupervised} uses representation learnt by a discriminator. Another line of research extends GAN to bidirectional framework (BiGAN) by introducing an auxiliary encoder branch that predicts the latent vector from a natural image \cite{donahue2016adversarial,donahue2019large}. The encoder learnt via BiGAN framework can be used as feature extractor for downstream tasks \cite{donahue2019large}. The use of GANs as universal feature extractors has severe limitations. First, GANs are not always capable of learning a multimodal distribution as they tend to suffer from mode collapse \cite{liu2019spectral}. The trade-off between GAN precision and recall is still difficult to control \cite{kynkaanniemi2019improved}. Besides, training a GAN on a large dataset of high-resolution images requires an extremely large computational budget, which makes ImageNet-scale experiments prohibitively expensive. Our approach differs from this line of work, as we use a GAN to specialize a model to a particular domain rather than trying to obtain universal feature representations. We explore the representation of a GAN generator that, to the best of our knowledge, has not been previously considered for transfer learning.

Bau \etal \cite{bau2018gan} show that activations of a generator are highly correlated with semantic segmentation masks for the generated image. One of the means for analysis of latent space and internal representation of a generator is latent embedding, \ie finding a latent vector that corresponds to a particular image. Several methods for embedding images into GAN latent space have been proposed \cite{karras2019analyzing,bau2019seeing,abdal2019image2stylegan}, bringing interesting insights about generator representations. For instance, it allowed to demonstrate that some of semantic categories are missing systematically in GAN-generated images \cite{bau2019seeing}. Similarly to these works we invert a GAN generator using both feature approximation and image reconstruction losses, although we do not aim at reconstructing the latent code, and only approximate the activations of the layers of a generator.

While image-level classification has been extensively studied in a semi-supervised setting, dense pixel-level classification with limited data has only drawn attention recently. Most of the works on semi-supervised semantic segmentation borrow the ideas from semi-supervised image classification and generalize them on high-resolution tasks. \cite{hung2018adversarial} adopt an adversarial learning scheme and propose a fully convolutional
discriminator that learns to differentiate ground truth label maps from probability maps
of segmentation predictions. \cite{mittal2019semi} use two network branches that link semi-supervised classification with semi-supervised segmentation including self-training.



\section{Conclusion}
We study the use of GAN generators for the task of learning domain-specific representations. We show that the representation of a GAN generator can be easily projected onto semantic segmentation map using a lightweight decoder. Then, we propose LayerMatch scheme for unsupervised domain-specific pretraining that is based on approximating  the generator representation. We present experiments in semi-supervised learning scenario and compare to recent semi-supervised semantic segmentation methods.

\clearpage
\small
\bibliographystyle{apalike}
\bibliography{bibliography}

\begin{thebibliography}{}

\bibitem[Abdal et~al., 2019]{abdal2019image2stylegan}
Abdal, R., Qin, Y., and Wonka, P. (2019).
\newblock Image2stylegan: How to embed images into the stylegan latent space?
\newblock In {\em Proceedings of the IEEE International Conference on Computer
  Vision}, pages 4432--4441.

\bibitem[Arjovsky et~al., 2017]{arjovsky2017wasserstein}
Arjovsky, M., Chintala, S., and Bottou, L. (2017).
\newblock Wasserstein gan.
\newblock {\em arXiv preprint arXiv:1701.07875}.

\bibitem[Bau et~al., 2018]{bau2018gan}
Bau, D., Zhu, J.-Y., Strobelt, H., Zhou, B., Tenenbaum, J.~B., Freeman, W.~T.,
  and Torralba, A. (2018).
\newblock Gan dissection: Visualizing and understanding generative adversarial
  networks.
\newblock {\em arXiv preprint arXiv:1811.10597}.

\bibitem[Bau et~al., 2019]{bau2019seeing}
Bau, D., Zhu, J.-Y., Wulff, J., Peebles, W., Strobelt, H., Zhou, B., and
  Torralba, A. (2019).
\newblock Seeing what a gan cannot generate.
\newblock In {\em Proceedings of the IEEE International Conference on Computer
  Vision}, pages 4502--4511.

\bibitem[Brock et~al., 2018]{brock2018large}
Brock, A., Donahue, J., and Simonyan, K. (2018).
\newblock Large scale gan training for high fidelity natural image synthesis.
\newblock {\em arXiv preprint arXiv:1809.11096}.

\bibitem[Chen et~al., 2018]{deeplabv3plus2018}
Chen, L.-C., Zhu, Y., Papandreou, G., Schroff, F., and Adam, H. (2018).
\newblock Encoder-decoder with atrous separable convolution for semantic image
  segmentation.
\newblock In {\em ECCV}.

\bibitem[Chen et~al., 2016]{chen2016infogan}
Chen, X., Duan, Y., Houthooft, R., Schulman, J., Sutskever, I., and Abbeel, P.
  (2016).
\newblock Infogan: Interpretable representation learning by information
  maximizing generative adversarial nets.
\newblock In {\em Advances in neural information processing systems}, pages
  2172--2180.

\bibitem[Donahue et~al., 2016]{donahue2016adversarial}
Donahue, J., Kr{\"a}henb{\"u}hl, P., and Darrell, T. (2016).
\newblock Adversarial feature learning.
\newblock {\em arXiv preprint arXiv:1605.09782}.

\bibitem[Donahue and Simonyan, 2019]{donahue2019large}
Donahue, J. and Simonyan, K. (2019).
\newblock Large scale adversarial representation learning.
\newblock In {\em Advances in Neural Information Processing Systems}, pages
  10541--10551.

\bibitem[Goodfellow et~al., 2014]{goodfellow2014generative}
Goodfellow, I., Pouget-Abadie, J., Mirza, M., Xu, B., Warde-Farley, D., Ozair,
  S., Courville, A., and Bengio, Y. (2014).
\newblock Generative adversarial nets.
\newblock In {\em Advances in neural information processing systems}, pages
  2672--2680.

\bibitem[Hung et~al., 2018]{hung2018adversarial}
Hung, W.-C., Tsai, Y.-H., Liou, Y.-T., Lin, Y.-Y., and Yang, M.-H. (2018).
\newblock Adversarial learning for semi-supervised semantic segmentation.
\newblock {\em arXiv preprint arXiv:1802.07934}.

\bibitem[Karras et~al., 2017]{karras2017progressive}
Karras, T., Aila, T., Laine, S., and Lehtinen, J. (2017).
\newblock Progressive growing of gans for improved quality, stability, and
  variation.
\newblock {\em arXiv preprint arXiv:1710.10196}.

\bibitem[Karras et~al., 2018]{karras2018style}
Karras, T., Laine, S., and Aila, T. (2018).
\newblock A style-based generator architecture for generative adversarial
  networks.
\newblock {\em arXiv preprint arXiv:1812.04948}.

\bibitem[Karras et~al., 2019a]{karras2019style}
Karras, T., Laine, S., and Aila, T. (2019a).
\newblock A style-based generator architecture for generative adversarial
  networks.
\newblock In {\em Proceedings of the IEEE Conference on Computer Vision and
  Pattern Recognition}, pages 4401--4410.

\bibitem[Karras et~al., 2019b]{karras2019analyzing}
Karras, T., Laine, S., Aittala, M., Hellsten, J., Lehtinen, J., and Aila, T.
  (2019b).
\newblock Analyzing and improving the image quality of stylegan.
\newblock {\em arXiv preprint arXiv:1912.04958}.

\bibitem[Kingma and Dhariwal, 2018]{kingma2018glow}
Kingma, D.~P. and Dhariwal, P. (2018).
\newblock Glow: Generative flow with invertible 1x1 convolutions.
\newblock In {\em Advances in Neural Information Processing Systems}, pages
  10215--10224.

\bibitem[Kynk{\"a}{\"a}nniemi et~al., 2019]{kynkaanniemi2019improved}
Kynk{\"a}{\"a}nniemi, T., Karras, T., Laine, S., Lehtinen, J., and Aila, T.
  (2019).
\newblock Improved precision and recall metric for assessing generative models.
\newblock In {\em Advances in Neural Information Processing Systems}, pages
  3929--3938.

\bibitem[Larsen et~al., 2015]{larsen2015autoencoding}
Larsen, A. B.~L., S{\o}nderby, S.~K., Larochelle, H., and Winther, O. (2015).
\newblock Autoencoding beyond pixels using a learned similarity metric.
\newblock {\em arXiv preprint arXiv:1512.09300}.

\bibitem[Lee et~al., 2019]{lee2019maskgan}
Lee, C.-H., Liu, Z., Wu, L., and Luo, P. (2019).
\newblock Maskgan: towards diverse and interactive facial image manipulation.
\newblock {\em arXiv preprint arXiv:1907.11922}.

\bibitem[Lee, 2013]{lee2013pseudo}
Lee, D.-H. (2013).
\newblock Pseudo-label: The simple and efficient semi-supervised learning
  method for deep neural networks.
\newblock In {\em Workshop on challenges in representation learning, ICML},
  volume~3, page~2.

\bibitem[Liu et~al., 2019]{liu2019spectral}
Liu, K., Tang, W., Zhou, F., and Qiu, G. (2019).
\newblock Spectral regularization for combating mode collapse in gans.
\newblock In {\em Proceedings of the IEEE International Conference on Computer
  Vision}, pages 6382--6390.

\bibitem[Liu et~al., 2015]{liu2015faceattributes}
Liu, Z., Luo, P., Wang, X., and Tang, X. (2015).
\newblock Deep learning face attributes in the wild.
\newblock In {\em Proceedings of International Conference on Computer Vision
  (ICCV)}.

\bibitem[Makhzani et~al., 2015]{makhzani2015adversarial}
Makhzani, A., Shlens, J., Jaitly, N., Goodfellow, I., and Frey, B. (2015).
\newblock Adversarial autoencoders.
\newblock {\em arXiv preprint arXiv:1511.05644}.

\bibitem[Mescheder et~al., 2018]{mescheder2018training}
Mescheder, L., Geiger, A., and Nowozin, S. (2018).
\newblock Which training methods for gans do actually converge?
\newblock {\em arXiv preprint arXiv:1801.04406}.

\bibitem[Mittal et~al., 2019]{mittal2019semi}
Mittal, S., Tatarchenko, M., and Brox, T. (2019).
\newblock Semi-supervised semantic segmentation with high-and low-level
  consistency.
\newblock {\em IEEE Transactions on Pattern Analysis and Machine Intelligence}.

\bibitem[Miyato et~al., 2018]{miyato2018spectral}
Miyato, T., Kataoka, T., Koyama, M., and Yoshida, Y. (2018).
\newblock Spectral normalization for generative adversarial networks.
\newblock {\em arXiv preprint arXiv:1802.05957}.

\bibitem[Radford et~al., 2015]{radford2015unsupervised}
Radford, A., Metz, L., and Chintala, S. (2015).
\newblock Unsupervised representation learning with deep convolutional
  generative adversarial networks.
\newblock {\em arXiv preprint arXiv:1511.06434}.

\bibitem[Sun et~al., 2019]{sun2019high}
Sun, K., Zhao, Y., Jiang, B., Cheng, T., Xiao, B., Liu, D., Mu, Y., Wang, X.,
  Liu, W., and Wang, J. (2019).
\newblock High-resolution representations for labeling pixels and regions.
\newblock {\em arXiv preprint arXiv:1904.04514}.

\bibitem[Xiao et~al., 2018]{xiao2018unified}
Xiao, T., Liu, Y., Zhou, B., Jiang, Y., and Sun, J. (2018).
\newblock Unified perceptual parsing for scene understanding.
\newblock In {\em Proceedings of the European Conference on Computer Vision
  (ECCV)}, pages 418--434.

\bibitem[Yu et~al., 2015a]{yu2015lsun}
Yu, F., Seff, A., Zhang, Y., Song, S., Funkhouser, T., and Xiao, J. (2015a).
\newblock Lsun: Construction of a large-scale image dataset using deep learning
  with humans in the loop.
\newblock {\em arXiv preprint arXiv:1506.03365}.

\bibitem[Yu et~al., 2015b]{yu15lsun}
Yu, F., Zhang, Y., Song, S., Seff, A., and Xiao, J. (2015b).
\newblock Lsun: Construction of a large-scale image dataset using deep learning
  with humans in the loop.
\newblock {\em arXiv preprint arXiv:1506.03365}.

\bibitem[Zhang et~al., 2016]{zhang2016colorful}
Zhang, R., Isola, P., and Efros, A.~A. (2016).
\newblock Colorful image colorization.
\newblock In {\em European conference on computer vision}, pages 649--666.
  Springer.

\end{thebibliography}
\end{document}